\documentclass[10pt,journal]{IEEEtran}
\usepackage{amsmath,amssymb,booktabs,graphicx,multirow,xcolor,url}
\usepackage[caption=false,font=footnotesize]{subfig}
\usepackage{microtype}
\graphicspath{{./}}
\title{Artifact Annotations Partially Substitute for Per-User Calibration: SAFE-EDA and a Normalization-Controlled Evaluation of Wrist-EDA Affect Recognition}
\author{Haochen Chai, Xinbi Luo, Zining Liu, and Fangfang Jiang\textsuperscript{*}\thanks{This work was supported in part by the Fundamental Research Funds for the Central Universities under Grant N26GFZ002 and in part by the National Natural Science Foundation of China under Grant 62427807. Haochen Chai, Xinbi Luo, Zining Liu, and Fangfang Jiang are with the College of Medicine and Biological Information Engineering, Northeastern University, Shenyang, China (e-mail: chaihc@mails.neu.edu.cn; 20237351@stu.edu.cn; liuzn1@mails.neu.edu.cn; jiangff@bmie.neu.edu.cn). Haochen Chai: ORCID 0009-0007-4785-2912; Xinbi Luo: ORCID 0009-0002-3406-884X; Zining Liu: ORCID 0009-0002-5403-0478. *Fangfang Jiang is the corresponding author.}}
\begin{document}
\bstctlcite{IEEEStyleControl}
\raggedbottom
\maketitle

\begin{abstract}Wrist electrodermal activity (EDA) differs in amplitude from one person to the next, so affect-recognition models normalize their input before classification. Studies that test such models on held-out subjects seldom report where the normalization statistics come from, yet statistics computed from the held-out subject's own recording give the model information that a device does not have when it is first worn. We asked how this choice alters the measured benefit of pretraining. A compact convolutional network, SAFE-EDA, was pretrained on expert artifact annotations from 43 subjects and compared with the same network trained from scratch on the Wearable Stress and Affect Detection (WESAD) dataset (15 subjects, leave-one-subject-out), with two normalization sources crossed with four window hops. When the statistics came only from training subjects, pretraining raised macro-F1 by 0.078 to 0.227; when they came from the held-out user's full recording, the gain fell to between 0.020 and 0.050 and was no longer significant. Artifact supervision was far more useful than self-supervised pretraining on the same recordings (0.078 versus 0.008). Across 13 configurations in two datasets, the pretrained network was better in 12, but on the second dataset (26 subjects) per-user normalization increased the gain instead of reducing it, so the interaction depends on the data. Only five of 50 published WESAD studies state which data were used for normalization. Reporting this choice is necessary to separate first-use performance from performance after calibration.
\end{abstract}

\section{Introduction}
Wearable sensors make it possible to follow stress and affect outside the laboratory, but a prediction can only be interpreted together with the information that was available when it was made. Electrodermal activity (EDA) illustrates the problem well. Its amplitude reflects sympathetic arousal, and also skin properties, sensor contact, and other individual and recording factors~\cite{boucsein2012eda}, so EDA models usually normalize the signal before classification. A device worn for the first time has to do this with statistics learned from other people. A retrospective benchmark can instead use statistics computed from the new user's complete recording, a procedure we refer to as per-user calibration. Both procedures can be run with the same sensor and the same trained classifier, but only the first tells us how the model performs at first use; a study that uses per-user calibration evaluates a model that has already seen the scale of the person it classifies. Neither procedure is wrong, and in longitudinal monitoring a short calibration period may be perfectly acceptable. The difficulty arises when a reader cannot tell which of the two was used.

Existing reporting guidance asks for this kind of information in general terms. TRIPOD+AI and the original TRIPOD statement relate the performance of a prediction model to the population, inputs, development procedure, and evaluation setting that produced it~\cite{collins2024tripodai,collins2015tripod}; PROBAST and MI-CLAIM turn the same concern into risk-of-bias assessment and minimum reporting items for clinical artificial intelligence~\cite{wolff2019probast,norgeot2020miclaim}, and CONSORT-AI and SPIRIT-AI extend it to trials and protocols~\cite{liu2020consortai,cruzrivera2020spiritai}. Work on leakage shows that information can cross an apparently valid train/test split~\cite{kapoor2023leakage,kaufman2012leakage}, and work on cross-validation shows that choosing a model on the evaluated data biases its reported error~\cite{varma2006bias,cawley2010overfitting}. In wearable stress recognition, the Wearable Stress and Affect Detection (WESAD) dataset has become the common benchmark~\cite{schmidt2018wesad}, and comparisons of personalized and participant-exclusive evaluation on consumer devices~\cite{li2024personalized}, as well as reviews of generalization~\cite{vos2023generalizable}, show that estimates change considerably with the protocol. What these sources leave open is the status of unlabeled data from the evaluated subject. A subject-independent split keeps that person's labels out of training, but it says nothing about whether their unlabeled signal was used to set the input scale. When we read the Methods sections of 50 published WESAD studies, five stated where the normalization statistics came from (10\%, 95\% Wilson interval 4--21\%), whereas 26 stated the window length (52\%, 39--65\%).

The omission matters for more than bookkeeping, because normalization and pretraining may address the same difficulty. If a pretrained network has learned to be robust to differences in signal amplitude, normalizing with the user's own statistics removes part of the variation it was trained to handle, and the measured benefit of pretraining should shrink. The value of pretraining reported in a study would then depend on a preprocessing choice that the study may not mention. A second protocol choice, the hop between successive windows, can interact with the first. Dense windowing multiplies the number of training examples without adding independent participants, and a network trained from scratch may depend on this redundancy more than a pretrained one.

We examined these questions with SAFE-EDA, a compact convolutional network whose temporal layers are pretrained on expert, sample-level artifact annotations from the EDABE dataset and then fine-tuned for affect recognition. The pretrained network and an identical network trained from scratch share the architecture, subject folds, training budget, and evaluation rules, so that they differ only in their initialization. On WESAD, we crossed two normalization sources, training subjects only or the held-out user's full recording, with four window hops, and based inference on subject-paired contrasts. A second public dataset of 26 subjects tested the same two factors in separate scans. To find out what the pretraining contributes, we compared artifact supervision with self-supervised pretraining on the same recordings and with pretraining on shuffled artifact labels, and we ablated components of the backbone.

The study makes three contributions. First, it shows that on WESAD the benefit of artifact-supervised pretraining depends strongly on the normalization source: the gain over training from scratch ranged from 0.020 to 0.227, and it was largest when the model had to rely on statistics from training subjects and when windows overlapped less. On the second dataset the dependence reversed in direction, which we report as part of the result. Second, the supervision controls show that the expert labels matter: pretraining on artifact labels gave a gain about ten times larger than self-supervision on the same recordings, and correct labels added to what shuffled labels achieved. Third, the literature survey identifies the source of normalization statistics as a concrete reporting item, and the experiment shows what a reader cannot judge when it is missing.

\begin{figure*}[t]
\centering\includegraphics[width=\textwidth]{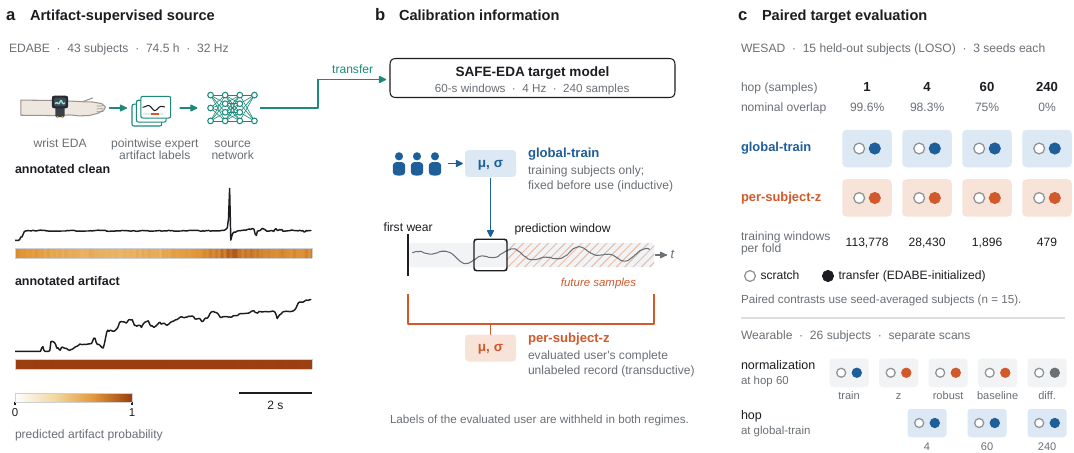}
\caption{SAFE-EDA system and evaluation. (a) Pointwise expert artifact labels in EDABE supervise a source network whose stem and residual blocks are transferred to the target model. Traces are 8-s, 32-Hz EDABE examples selected by annotation and plotted on independent vertical scales; the strip beneath each trace shows the source-head artifact probability on a common zero-to-one scale. (b) Global-train normalization uses statistics fixed on training subjects and supports inductive use; per-subject-z computes statistics from the evaluated user's complete unlabeled recording, including samples after the prediction window. (c) The WESAD grid crosses two normalization sources with four hops, and every cell contains both initialization arms; training windows per fold are listed below the grid. The Wearable dataset uses a five-transform normalization ladder at hop 60 and a three-hop scan under global-train normalization.}\label{fig:calibration}
\end{figure*}

\section{Related Work}
EDA processing has a long methodological literature. Reviews describe how acquisition and processing choices shape the meaning of derived measures~\cite{posada2020eda_review}, and toolboxes such as NeuroKit2 and pyEDA have made common operations reproducible~\cite{makowski2021neurokit2,aqajari2021pyeda}. Decomposition methods separate slow tonic activity from phasic responses under explicit model assumptions~\cite{benedek2010decomposition,greco2016cvxeda}. Motion artifacts are a well-known source of error in wrist recordings, and several studies have proposed ways to detect or correct them~\cite{hossain2022motioneda,hossain2021preliminary,zhang2017unsupervisededa}. EDABE contributed expert annotations for individual samples, and later work has treated signal quality as a learning target in its own right~\cite{llanes2023edabe,kong2024uneteda}. Our source task builds on these resources with a different aim. We do not use artifact detection to clean the signal; we ask whether a network that has learned to recognize artifacts is a better starting point for recognizing affect. When performance on WESAD is compared across studies, it is also easy to overlook that the reported scores often rest on different inputs. The original benchmark combined several chest and wrist modalities~\cite{schmidt2018wesad}, context-aware fusion adds contextual information~\cite{relatedW005}, and Ahmad and Khan classify from ECG~\cite{ahmad2025vit}. We therefore use published scores only to indicate what different sensor configurations achieve and compare methods within our own protocol; the architectures used in the surveyed studies are listed in the supplementary coding table.

The boundary between training and evaluation information is most visible in work on generalization and personalization. Li and Washington compared personalized, participant-inclusive, and participant-exclusive models~\cite{li2024personalized}, Vos et al. reviewed why wearable stress models generalize poorly~\cite{vos2023generalizable}, and cross-dataset and transfer studies have drawn attention to differences in population and acquisition~\cite{benchekroun2023crossdataset,albaladejo2022configurations,berkemeier2025affective}. In this literature, the label ``subject-independent'' covers designs that differ in an important respect, because a model may never see the labels of the evaluated person and still use that person's unlabeled measurements to set the feature scale. Published descriptions show how varied the choices are. Mitro et al. scale features with means and standard deviations from the training samples~\cite{relatedW025}, Albaladejo-Gonz\'{a}lez et al. normalize heart rate against an individual baseline~\cite{albaladejo2022configurations}, Yang et al. remove a subject-specific baseline~\cite{relatedW041}, and Liapis et al. add categorical embeddings that change what the classifier receives~\cite{relatedW015}. These studies concern different signals and tasks, and we cite them to show the range of practice rather than as instances of our two conditions. The timing of the information differs as well: a baseline recorded before use becomes available after a defined onboarding period, whereas standardization over the complete recording draws on samples recorded after the prediction is made, and neither the normalization formula nor the type of split reveals which of these applies. Oliver and Dakshit's comparison of modalities on WESAD makes a related point about stating what is being compared~\cite{relatedW060}. Individual differences are, of course, widely recognized. Our question is the narrower one of what changes when the source of the normalization statistics is varied on its own, with the architecture held fixed.

Methodological work on leakage provides the remaining background. Kapoor and Narayanan documented how failures of information separation have affected machine-learning-based science~\cite{kapoor2023leakage}, and Kaufman et al. gave a formal account of leakage that goes beyond the mere presence of a train/test split~\cite{kaufman2012leakage}. Both show that preprocessing and model selection can carry information across an evaluation boundary even when the target labels never cross it. Varma and Simon, and Cawley and Talbot, showed why model selection and error estimation must be kept apart~\cite{varma2006bias,cawley2010overfitting}, and dependence-aware cross-validation addresses the difference between the number of observations and the number of independent units~\cite{roberts2017crossvalidation}. Reproducibility studies in machine learning and in physiological stress detection have given these concerns concrete form~\cite{pineau2021reproducibility,gjoreski2020reproducibility}. The point about independent units applies directly to our design. Overlapping windows and repeated random seeds produce many observations per subject without adding a participant, so we average seeds within each subject, test subject-paired contrasts, and keep random-window results only as a reference for how much leakage can inflate a score. We do not regard the use of a subject's unlabeled recording as label leakage. It is, however, an information condition that clinical prediction guidelines would expect authors to report~\cite{collins2024tripodai,wolff2019probast}, and this paper approaches it from two sides: the survey records what published studies say about it, and the experiment varies it within a single implementation.

\section{Materials and Methods}
\subsection{Study design}
The experiment was designed to isolate the effect of initialization while the source of normalization statistics was varied as an independent factor. Studies that change the network and the preprocessing at the same time cannot separate the two, so we fixed the target backbone and training budget and compared initializations within the same participant folds (Fig.~\ref{fig:calibration}). Three pretraining objectives on the same source recordings, namely correct artifact labels, shuffled labels, and self-supervision, allow the effect of annotation content to be distinguished from that of pretraining as such, and ablations of the backbone trained from scratch indicate how much the architecture alone contributes. We use the term partial substitution in an operational sense: two sources of information partially substitute for each other if the gain from one becomes smaller once the other is present. The definition refers to measured gains and makes no assumption about the physiological mechanism involved.

\subsection{Methods-text survey}
We drew a fixed, structured sample of WESAD studies and removed duplicates by DOI. Protocol fields were coded from the Methods section or an equivalent description of the experimental setup, and a field counted as reported only when the text specified it completely. Naming a scaler without stating which data it was fitted on, for example, did not establish the source of normalization statistics, and a train/test ratio without its allocation unit did not establish the split. Each field was counted once per study if any eligible part of the text specified it. Every code is accompanied by the section-level location and an excerpt of the source text, and the machine-readable supplement gives the quotation for each field and the locations searched when a field could not be resolved. One author read the Methods text of every study and checked each coded field against it. Study-level codes are given in the Supplementary Material. Proportions are reported with 95\% Wilson score intervals and describe the retrieved sample.

\subsection{Datasets and Cohorts}
EDABE provides 74.5 h of EDA recorded at 32 Hz from 43 subjects, with expert artifact annotations for every sample, and serves as the source dataset. WESAD provides wrist recordings from 15 subjects labeled as baseline, stress, or amusement. The main WESAD analysis uses 15-fold leave-one-subject-out (LOSO) evaluation, resampling to 4 Hz, 60-s windows, and hops of 1, 4, 60, or 240 samples. From the PhysioNet Wearable Device Dataset v1.0.0 (hereafter Wearable), we kept the 26 identifiers with readable EDA.csv files in all three sessions (AEROBIC, ANAEROBIC, and STRESS). Of the 36 candidate identifiers, six (S12 and f14--f18) have no public AEROBIC directory, consistent with the activity counts reported for the release, and two (f12 and f13) are complete in the public archive but were missing from the processed copy we used. All comparisons are paired within subject. The three datasets are public, and the study is a secondary analysis: EDABE v2 (Mendeley Data, DOI 10.17632/w8fxrg4pv5.2) was obtained from the release that accompanies the artifact-recognition study~\cite{llanes2023edabe}, WESAD from its 2018 release (DOI 10.1145/3242969.3242985)~\cite{schmidt2018wesad}, and Wearable from the PhysioNet v1.0.0 directory (\url{https://physionet.org/files/wearable-device-dataset/1.0.0/}).

\subsection{Normalization Regimes and Model}
The two main normalization regimes differ only in where the statistics come from. Global-train fits the mean and standard deviation on the training subjects and applies them unchanged to the held-out subject, which corresponds to inductive use. Per-subject-z estimates them from all unlabeled samples of the held-out subject; it is transductive, and it is how we implement per-user calibration. In both regimes, the labels of the held-out subject are withheld until scoring. Robust, baseline-referenced, and first-difference transforms serve as sensitivity analyses. Because per-subject-z uses the complete recording, including samples recorded after a given classification window (Fig.~\ref{fig:calibration}), it corresponds to retrospective evaluation. A causal deployment would rely on training statistics or on a defined segment recorded before use; the minimum duration of such a segment and the effect of online updating remain to be studied.

\begin{figure*}[t]
\centering\includegraphics[width=\textwidth]{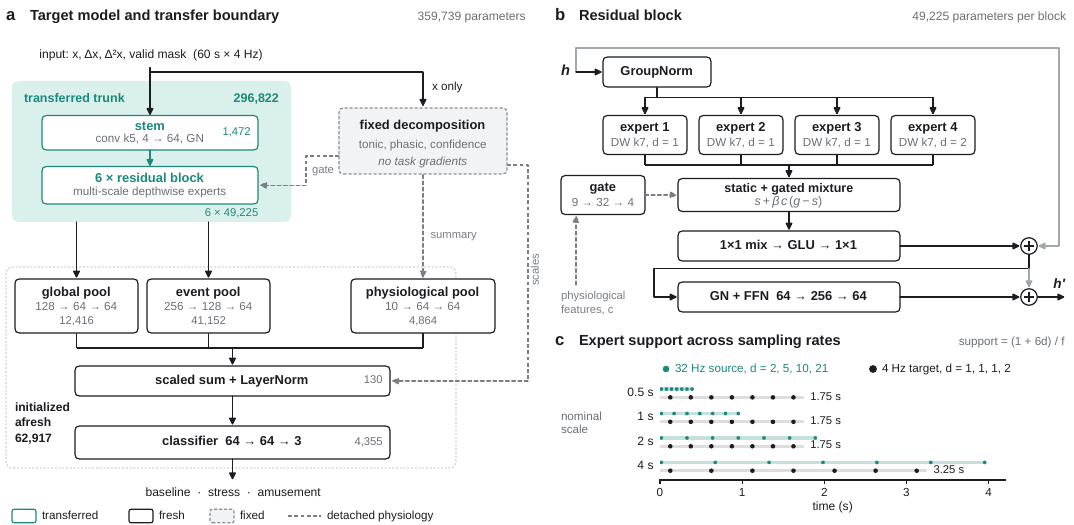}
\caption{SAFE-EDA architecture, with parameter counts taken from the model definition. (a) The stem and six residual blocks transfer 296,822 parameters; pooling, fusion, and classification initialize 62,917 parameters afresh (359,739 in total). The all-valid input channel is $1$, and only $x$ enters the fixed decomposition. Dashed gray paths carry detached physiological information. Global and event pooling use hidden features, whereas physiological pooling uses the decomposition summary. Pool boxes give MLP widths, with parameter counts below. The stem has 1,472 parameters, fusion 130, and the classifier 4,355. (b) Four parallel width-7 depthwise experts are combined with weights $s+\beta c(g-s)$, where $s$ is the static softmax, $g$ the physiological-feature gate, $c$ the reconstruction confidence, and $\beta$ a learned sigmoid scale. Channel mixing and the feed-forward network each have a residual connection; dropout and stochastic depth are omitted for clarity. (c) Kernel taps of each expert at the 32-Hz source rate (teal) and the 4-Hz target rate (black). The dilation is $d=\max\{1,\operatorname{round}[(r f_s-1)/(k-1)]\}$, and labels give the target support $(1+6d)/f_s$; the three shortest target experts share a 1.75-s support. Supports refer to single experts, not to the full network. The source artifact head is discarded before transfer.}\label{fig:model}
\end{figure*}

SAFE-EDA is a multi-scale convolutional network built from dilated depthwise-separable layers with confidence gating, with 359,739 parameters in total. The 296,822 parameters of the stem (`stem.*') and the residual blocks (`blocks.*') are transferred from the source model, and the target classification head is reinitialized in every fold (Fig.~\ref{fig:model}). Processing one 60-s window takes 140.6 MFLOPs, and the measured single-window latency on a CPU is 9.97 ms. A fixed decomposition branch plays a secondary role; at the sparsity setting used for deployment, its driver is exactly zero in 93\% of EDABE, 81\% of WESAD, and 82\% of Wearable windows.

\subsection{Signal representation and transfer boundary}
Source and target windows differ in both sampling rate and duration. A source window contains 256 samples at 32 Hz (8 s) and a target window 240 samples at 4 Hz (60 s). What carries over between the tasks are the weights of the temporal stem and blocks; source examples are not stretched to the target duration. The stem receives four channels, the normalized signal, its first and second differences, and a mask marking valid samples, and maps them to 64 channels with a convolution of width 5, followed by group normalization and a nonlinear activation. Each of the six residual blocks combines four depthwise convolutions of width 7 with a gate driven by physiological features and with a feed-forward stage that expands the 64 channels to 256 and projects them back. During source training, a pointwise head predicts artifacts; this head is not transferred, and the pooling and classification layers of the target model are initialized afresh.

The four experts in each block are intended to cover 0.5, 1, 2, and 4 s. For a nominal duration $r$, sample rate $f_s$, and kernel width $k=7$, the implementation sets the dilation to $d=\max\{1,\operatorname{round}[(r f_s-1)/(k-1)]\}$, which gives dilations of 2, 5, 10, and 21 at the source rate and of 1, 1, 1, and 2 at the target rate. Because the dilation cannot fall below one, the three shorter experts have the same support at 4 Hz, $(1+6d)/f_s=1.75$ s, and only the longest reaches 3.25 s, so the nominal durations are only approximated in the target task. These values refer to a single expert; the receptive field of the stacked network is larger. Keeping the kernel width fixed in this way leaves the parameter shapes unchanged across sampling rates, which is what allows the weights to be transferred.

\subsection{Fixed decomposition and diagnostic confidence}
The decomposition branch receives no gradients from the target task. A fixed smooth projection estimates tonic activity, and an iterative nonnegative sparse driver generates the phasic component through a fixed response kernel. Their sum is the reconstruction, and its difference from the input is the residual. A smoothed absolute residual defines a time-varying confidence whose temporal mean enters the pooling path, so the confidence is a deterministic measure of how well the decomposition reproduces the input. It is neither a probability that a classification is correct nor an artifact probability, which only the source head provides, and only on the source task. The decomposition parameters were fixed without reference to target labels.

Saved examples show what these operations produce. The EDABE examples were chosen by their annotated artifact fraction and the WESAD examples by the highest and lowest checkpoint confidence in the saved diagnostic collection; both WESAD examples come from subject S2. For display, the input normalization was reversed with the stored mean and standard deviation, so that tonic and reconstruction traces regain the offset, phasic, driver, and residual traces are rescaled without it, and all traces keep the scale of the source signal. Driver traces that are zero are shown as such, since an inactive driver is itself a property of the computation. The reconstruction is a model estimate, and no clean reference waveform exists for these recordings against which it could be checked.

\begin{figure}[t]
\centering\includegraphics[width=\columnwidth]{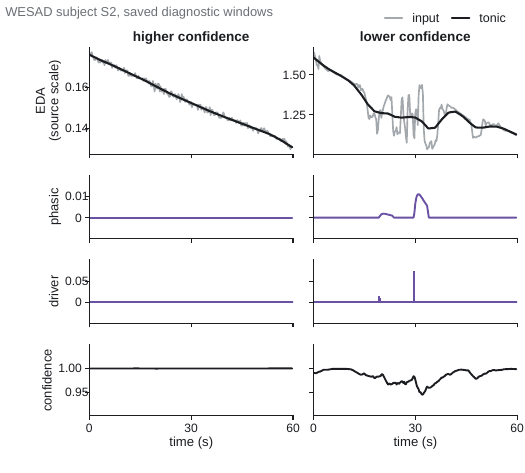}
\caption{Saved WESAD S2 windows with higher and lower checkpoint confidence: input and tonic traces, phasic response, sparse driver, and reconstruction confidence. Time is in seconds, and amplitudes use the exported source scale. Phasic, driver, and confidence axes are shared across columns; input axes differ to preserve waveform detail. The two windows are illustrative extremes, and confidence reflects reconstruction quality rather than the probability that a classification is correct.}\label{fig:mechanism}
\end{figure}

\subsection{Protocol axes and evaluation controls}
Three protocol factors were varied separately: the evaluation split, the normalization source, and the window hop. Random assignment of windows after segmentation is included as a reference that is known to be sensitive to leakage, and a fixed holdout of one subject-sized recording shows what a single partition yields. Leave-one-subject-out, which places every subject in the evaluation role once, is the design on which inference rests. At 4 Hz, a 60-s window contains 240 samples, so hops of 1, 4, 60, and 240 samples correspond to nominal overlaps of 99.6\%, 98.3\%, 75\%, and 0\%. The WESAD grid thus has eight cells, two normalization regimes by four hops, and each cell contains both arms, three seeds, and 15 held-out subjects. The source task uses the public pointwise artifact mask of EDABE, and the target head is initialized anew in every fold, in line with established practice for EDA preprocessing and artifact reporting~\cite{posada2020eda_review,makowski2021neurokit2,aqajari2021pyeda}. All cells were run with the same fixed budget and all are reported; no held-out label, probability, or checkpoint was used to choose among them.

For a contiguous eligible segment of $L$ samples, a window length $W$, and a hop $h$, the number of candidate windows is $\max\{0,1+\lfloor(L-W)/h\rfloor\}$. Counts are summed over segments and training subjects, and segment boundaries and label eligibility make the totals depart slightly from an exact inverse scaling with the hop. Nominal overlap is $100(1-h/W)$ for $h\leq W$. A smaller hop therefore increases the redundancy of the training set without changing the number of independent participants, and we report the measured window counts per fold alongside the formula so that the sampling density of each protocol is explicit.

\subsection{Training and Statistical Analysis}
Both arms were trained with AdamW at a learning rate of 0.001 for a fixed 12 epochs, without early stopping. No held-out value was used to choose a cell, checkpoint, seed, epoch, or hyperparameter, and training losses are reported to show how far each arm had converged within this budget. Results from the three seeds were averaged within each subject before inference, so that each contrast rests on 15 WESAD or 26 Wearable paired observations. The primary tests are Wilcoxon signed-rank tests, with Holm correction across the eight WESAD cells. Effect sizes and percentile intervals were computed from the subject-level summaries, the intervals from 20,000 bootstrap resamples of subjects with a fixed analysis seed. A subject counts as a win when the seed-averaged score of the transfer arm exceeds that of the scratch arm. Because neighboring windows overlap, results at the window level are reported only descriptively, and summaries across configurations, which share subjects, are read as indications of consistency rather than as independent tests.

Equivalence is summarized descriptively against a prespecified absolute margin of 0.05 macro-F1. Expected calibration error and the Brier score, which concern the calibration of predicted probabilities and not the normalization of the input, were computed at the subject level without post-hoc temperature or threshold fitting. The participant-level display aligns subject identifiers across arms within each regime and uses unrounded, seed-averaged scores with equal weight for every subject.

\section{Results}
\subsection{Reporting in published WESAD studies}
Just over half of the 50 studies described how their data were split, but very few described how the input was normalized. A split type could be identified in 28 studies (56\%, 95\% Wilson CI 42--69\%), 13 of which used LOSO (26\%, 16--40\%), and 26 stated the window length (52\%, 39--65\%). Fewer gave the hop (15 studies; 30\%, 19--44\%) or stated the overlap explicitly (six; 12\%, 6--24\%), and 11 specified how metrics were aggregated. Only five identified the data from which the normalization statistics were computed (10\%, 4--21\%), and four of these five used the evaluated user's own window or baseline. The counts record whether a field is described, not whether the underlying pipeline was sound.

The gap between window length and normalization source is more than a difference in reporting habits. Window length determines how much signal enters a single example, whereas the normalization source determines whose observations, and possibly which future observations, shape the numerical representation of that example, so a study can report the first carefully and still leave the second open. The intervals above describe each proportion separately; a formal comparison between fields would require their joint pattern across studies, which Supplementary Tables S1 and S2 provide.

\subsection{Effect of split and normalization}
Before turning to the full grid, we checked how strongly the evaluation split and the normalization source affect the scores of a single model (Table~\ref{tab:split}). With random assignment of windows, the transfer arm reached a macro-F1 of 0.9966, against 0.5794 under LOSO. The difference of 0.4172 arises because random assignment evaluates recordings that are already represented in training, and we use this result only as a reference for leakage. A single fixed holdout, with three seeds on one partition, reversed the order of the two arms (transfer 0.4787, scratch 0.5258), which shows how unstable estimates from one partition can be. All inference in the rest of the paper is therefore based on LOSO.

\begin{table}[t]
\caption{WESAD split comparison (mean macro-F1). Counts are scratch/transfer observations: 3/3 denotes three seeds of one partition, and 45/45 denotes 15 subjects with three seeds each. Only LOSO supports the subject-level paired test.}
\label{tab:split}\centering\scriptsize
\setlength{\tabcolsep}{2.5pt}
\begin{tabular}{@{}lrrrrr@{}}
\toprule Split & Scratch & Transfer & $n$ & Paired $p$ & Unit\\
\midrule Random window & 0.9508 & 0.9966 & 3/3 & -- & subject--seed\\
Fixed subject holdout & 0.5258 & 0.4787 & 3/3 & -- & subject--seed\\
Leave-one-subject-out & 0.5015 & 0.5794 & 45/45 & 0.00116 & subject--seed\\
\bottomrule
\end{tabular}
\end{table}

\begin{figure*}[t]
\centering
\includegraphics[width=\textwidth]{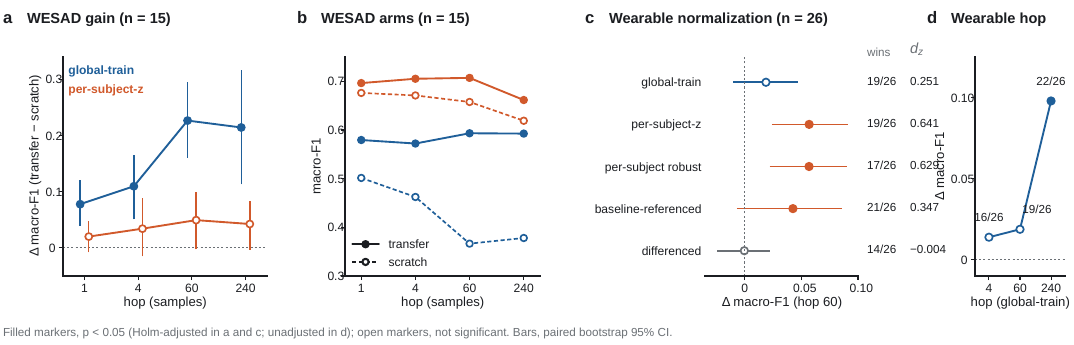}
\caption{Experiment atlas. (a) WESAD transfer gain (transfer $-$ scratch macro-F1) across hops under both normalization regimes, with paired bootstrap 95\% CIs (Data Table T17; $n=15$). (b) Mean macro-F1 of both arms. Hops are treated as categories. (c) Wearable normalization ladder at hop 60, with paired bootstrap 95\% CIs, wins, and Cohen's $d_z$ (Data Table T05; $n=26$). (d) Wearable hop scan under global-train normalization, with wins (Data Table T06; $n=26$). Filled markers denote $p<0.05$ (Holm-adjusted in a and c; unadjusted in d).}
\label{fig:atlas}
\end{figure*}

Under LOSO, the normalization source alone changed performance by a wide margin. Replacing training-subject statistics with statistics from the held-out subject raised scratch macro-F1 from 0.5015 to 0.6762 (+0.1747, $p=0.00537$) and transfer macro-F1 from 0.5794 to 0.6964 (+0.1170, $p=0.0181$). With per-user normalization and a hop of one sample, the transfer arm reached an accuracy of 0.7815 (95\% CI [0.7040,0.8485]), a macro-F1 of 0.6964, a balanced accuracy of 0.7127, and $\kappa=0.6376$ from wrist EDA alone. The original WESAD report gave an accuracy of 0.8034 with multimodal chest sensing, which lies within this interval. Because the protocols of the two reports are not fully aligned, the comparison speaks to what a single wrist sensor can achieve rather than to the merit of either method~\cite{schmidt2018wesad,ahmad2025vit}.

\subsection{The normalization--hop grid on WESAD}
The full grid shows that the benefit of pretraining depends on the normalization source (Table~\ref{tab:grid}, Fig.~\ref{fig:atlas}a). With global-train normalization, transfer improved on scratch at every hop, by 0.0779, 0.1098, 0.2268, and 0.2145 at hops of 1, 4, 60, and 240 samples, and all four gains remained significant after Holm correction. With per-subject-z normalization, the gains lay between 0.0202 and 0.0495, and none was significant. The difference between the two regimes, which is the interaction of interest, was largest at hop 60 (+0.1773, $p=0.000427$, 95\% CI [+0.1191,+0.2311]; Table~\ref{tab:interactionmain}).

The grid also shows where the larger gains under global-train normalization come from. As the hop increased from 1 to 60 samples, scratch macro-F1 fell by 0.1350, whereas the transfer arm changed by only 0.0139 (Fig.~\ref{fig:atlas}b). Even at hop 240, with fewer than 500 training windows per fold, the gain was still 0.2145. Under per-subject-z, the same increase in hop moved scratch from 0.6762 to 0.6576 and transfer from 0.6964 to 0.7071, and the gain changed by only 0.0293. Sparse windowing thus hurts the network trained from scratch mainly when its input scale has to be inferred from other subjects, and it is this loss that pretraining prevents. Conversely, dense windowing raises the scratch baseline against which pretraining is judged, so the measured value of pretraining depends on the hop as well as on the normalization. Since the hop is usually chosen with the computational budget of a device in mind, it deserves to be reported as carefully as the normalization source.

\begin{table*}[t]
\caption{WESAD normalization--hop grid ($n=15$ subjects per cell). Scores are mean macro-F1. Gains are given with paired 95\% CIs, and tests are Holm-adjusted. Overlap and windows per fold describe the protocol; losses are scratch/transfer training losses.}
\label{tab:grid}\centering\scriptsize
\setlength{\tabcolsep}{2.2pt}
\begin{tabular}{llrrr rrr l r l}
\toprule Regime & Hop & Overlap\% & Windows/fold & Scratch & Transfer & Gain [95\% CI] & Wins & $d_z$ & Holm $p$ & Losses\\
\midrule
Global-train & 1 & 99.6 & 113778 & 0.5015 & 0.5794 & +0.0779 [0.0394,0.1206] & 13/15 & 0.931 & 0.00812 & --\\
Global-train & 4 & 98.3 & 28430 & 0.4624 & 0.5722 & +0.1098 [0.0508,0.1655] & 13/15 & 0.932 & 0.02562 & 0.361/0.119\\
Global-train & 60 & 75.0 & 1896 & 0.3665 & 0.5933 & +0.2268 [0.1603,0.2951] & 15/15 & 1.642 & 0.00049 & 0.871/0.524\\
Global-train & 240 & 0.0 & 479 & 0.3781 & 0.5926 & +0.2145 [0.1130,0.3162] & 11/15 & 1.028 & 0.02685 & 0.935/0.641\\
Per-subject-z & 1 & 99.6 & 113778 & 0.6762 & 0.6964 & +0.0202 [-0.0078,0.0473] & 11/15 & 0.356 & 0.458 & --\\
Per-subject-z & 4 & 98.3 & 28430 & 0.6710 & 0.7053 & +0.0343 [-0.0151,0.0893] & 10/15 & 0.325 & 0.458 & 0.137/0.042\\
Per-subject-z & 60 & 75.0 & 1896 & 0.6576 & 0.7071 & +0.0495 [-0.0024,0.0995] & 10/15 & 0.478 & 0.2944 & 0.464/0.299\\
Per-subject-z & 240 & 0.0 & 479 & 0.6190 & 0.6616 & +0.0426 [-0.0029,0.0843] & 11/15 & 0.477 & 0.2944 & 0.560/0.451\\
\bottomrule
\end{tabular}
\end{table*}

\begin{table}[t]
\caption{Normalization--pretraining interaction (Data Table T18; $n=15$). The difference is the global-train gain minus the per-subject-z gain.}
\label{tab:interactionmain}\centering\scriptsize
\setlength{\tabcolsep}{2pt}
\begin{tabular}{rrrrrrr}
\toprule Hop & Global & Subject & Diff. & $p$ & 95\% CI & $n$\\
\midrule
1 & 0.0779 & 0.0202 & 0.0578 & 0.0637 & [0.0077,0.1096] & 15\\
4 & 0.1098 & 0.0343 & 0.0755 & 0.0353 & [0.0040,0.1447] & 15\\
60 & 0.2268 & 0.0495 & 0.1773 & 0.000427 & [0.1191,0.2311] & 15\\
240 & 0.2145 & 0.0426 & 0.1720 & 0.00061 & [0.0966,0.2499] & 15\\
\bottomrule
\end{tabular}
\end{table}

\begin{figure}[t]
\centering\includegraphics[width=\columnwidth]{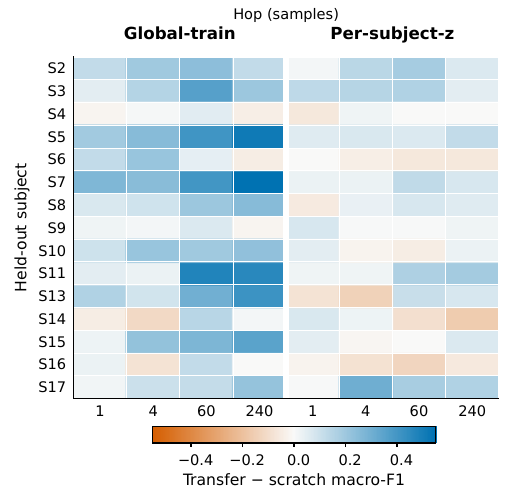}
\caption{Participant-level transfer effects in the WESAD normalization--hop grid (Data Table T21). Each cell is transfer minus scratch after averaging three seeds within a participant; rows keep the same 15 subject identifiers. The diverging color scale is shared across all eight conditions and centered at zero. Cell-wise and interaction tests are reported in Tables~\ref{tab:grid} and~\ref{tab:interactionmain}.}\label{fig:subjects}
\end{figure}

The cell means conceal considerable variation between participants (Fig.~\ref{fig:subjects}). With subject identifiers held fixed across columns, the display shows that the size of the effect differs from person to person and that, for some participants, its direction changes with the normalization source. The color scale is centered at zero and shared by both regimes, so equal intensity indicates an equal effect. Win counts add a further perspective: all 15 participants benefited from pretraining at global-train hop 60, whereas under per-subject-z the number of wins was 10 or 11 of 15.

\subsection{Stability across hops and configurations}
The pretrained network was also more stable across hops, but mainly under global-train normalization. Over the four hops, scratch macro-F1 varied by 0.1350 under global-train and by 0.0572 under per-subject-z, whereas the corresponding ranges for transfer were 0.0211 and 0.0455 (Data Table T19). The across-hop standard deviation of the scratch arm was 6.34 times that of the transfer arm under global-train, but only 1.22 times under per-subject-z. The training losses point in the same direction. At hop 60, they were 0.871 and 0.524 for scratch and transfer under global-train and 0.464 and 0.299 under per-subject-z, so within the same budget the pretrained network converged further, particularly when its input scale came from other subjects. Because training was fixed at 12 epochs, part of the difference between the arms therefore reflects the speed of optimization.

Seen from the other side, per-user normalization helped the network trained from scratch more than the pretrained one at every hop. Relative to global-train statistics, it raised scratch macro-F1 by 0.1747, 0.2087, 0.2911, and 0.2409 at hops 1, 4, 60, and 240, and transfer macro-F1 by 0.1170, 0.1331, 0.1138, and 0.0689 (Data Table T20). This asymmetry is what partial substitution predicts: if the user's own statistics supply part of the scale information that pretraining would otherwise provide, they should matter less to a pretrained network.

To summarize the direction of the effect across both datasets, we combined the normalization ladder with the nonduplicate global-train hop cells (Data Table T07). Transfer exceeded scratch in 6 of 6 WESAD and 6 of 7 Wearable configurations, 12 of 13 in total (one-sided sign test, nominal $p=0.00171$), and across these mixed settings the standard deviation of scratch was 2.61 and 2.88 times that of transfer. The configurations share subjects, so this summary indicates consistency rather than independent replication, and the regime-specific estimates above remain the basis for the stability result.

\subsection{What the pretraining contributes}
To find out what the pretrained network brings to the target task, we compared pretraining objectives at the dense-window setting under global-train normalization (Table~\ref{tab:controls}). The network trained from scratch reached a macro-F1 of 0.5015. Self-supervised pretraining on the same EDABE recordings, without their artifact annotations, reached 0.5090, a difference of 0.0075, whereas pretraining on the artifact labels reached 0.5794, a paired gain of 0.0779. Most of the benefit therefore comes from the supervised artifact objective rather than from exposure to the source recordings. The self-supervised and supervised arms differ in their objective as well as in label availability, however, so the comparison with shuffled labels is the more direct test of whether the content of the annotations matters.

\begin{table}[t]
\caption{Supervision and architecture controls at global-train, hop 1. F1 denotes macro-F1; gains and paired tests are relative to scratch. Deep-arm means from Data Table T14 summarize 15 subjects $\times$ 3 seeds, and tests use 15 seed-averaged subjects. Dashes denote unavailable tests. Shuffled F1 is 0.5015+0.0455; the self-supervised gain is 0.5090$-$0.5015.}
\label{tab:controls}\centering\scriptsize
\setlength{\tabcolsep}{2.2pt}
\begin{tabular}{@{}lrrrrl@{}}
\toprule Signal & F1 & $\Delta$ & Wins & $p$ & Source\\
\midrule None (scratch) & 0.5015 & -- & -- & -- & T14\\
Self-supervised & 0.5090 & +0.0075 & -- & -- & T14\\
Shuffled artifacts & 0.5470 & +0.0455 & 12/15 & 0.03015 & T08\\
True artifacts & 0.5794 & +0.0779 & 13/15 & 0.00116 & T14/T08\\
\midrule \multicolumn{4}{l}{Scratch architecture variant (T14)} & \multicolumn{2}{r}{F1}\\
\multicolumn{4}{l}{Static only} & \multicolumn{2}{r}{0.5005}\\
\multicolumn{4}{l}{Global pooling only} & \multicolumn{2}{r}{0.5006}\\
\multicolumn{4}{l}{Full SAFE-EDA} & \multicolumn{2}{r}{0.5015}\\
\multicolumn{4}{l}{Without decomposition/confidence} & \multicolumn{2}{r}{0.5094}\\
\bottomrule
\end{tabular}
\end{table}

Pretraining with shuffled labels reached 0.5470, between the self-supervised and the correctly labeled network, and correct labels remained clearly better (Section~IV-F). The architecture of the backbone, by contrast, made little difference. Variants trained from scratch lay between 0.5005 and 0.5094 (static experts only, 0.5005; global pooling only, 0.5006; the full network, 0.5015; without decomposition and confidence, 0.5094), and the full network was not the strongest of them. How much of the pretrained network was transferred mattered more: freezing the whole trunk gave 0.5380, transferring only the stem 0.5463, only the blocks 0.5507, and the full trunk 0.5794 (Data Table T14). Taken together, these comparisons identify artifact supervision as the useful ingredient of SAFE-EDA. A plausible explanation is that learning to recognize artifacts teaches the trunk to cope with differences in signal amplitude between people, although we did not measure this directly.

\begin{figure}[t]
\centering\includegraphics[width=\columnwidth]{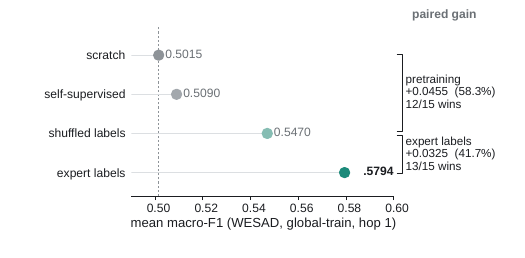}
\caption{Supervision ladder at WESAD global-train, hop 1 (Data Tables T14 and T08). Points are mean macro-F1 for scratch, self-supervised, shuffled-label, and expert-label initialization; the shuffled-label mean is the scratch mean plus the paired shuffled contrast. Brackets give the paired permutation contrasts: supervised pretraining with shuffled labels (+0.0455, 58.3\% of the total gain) and the further contribution of expert labels (+0.0325, 41.7\%), with subject wins. The self-supervised arm has no paired test. The ladder compares discrete source objectives, not doses of supervision.}\label{fig:supervision}
\end{figure}

Fig.~\ref{fig:supervision} places the four initializations on one axis. The step from scratch to self-supervision changes only the use of the source recordings, while the steps through shuffled and correct labels add first supervision and then label content; correct labels account for part of the gain and supervised pretraining as such for the rest. The two tasks also have different outputs, since a pointwise artifact mask is not an affect label. The transferred trunk can therefore carry useful signal structure while the target head learns its own decision rule, an interpretation that does not require the classifier to remove artifacts or to reconstruct a clean waveform.

\subsection{Specificity controls}
The source model did learn to detect artifacts. On the official EDABE test set, its artifact F1 was 0.5792, compared with 0.4234 for the released post-processed reference (+0.1558, 95\% CI [0.1196,0.1915], $d_z=2.522$; 10/10 wins, $p=0.00195$), and on the validation set the scores were 0.4654 and 0.3453 (+0.1201, 95\% CI [0.0471,0.2142], $d_z=0.987$; 6/7 wins, $p=0.03125$). These comparisons on the source recordings complement earlier work on artifact detection and correction~\cite{hossain2022motioneda,hossain2021preliminary,zhang2017unsupervisededa,kong2024uneteda}. The permutation control then divides the gain at global-train, hop 1, into two parts. Pretraining with shuffled labels retained 0.0455 of it (58.3\%; 12/15 wins, $p=0.03015$), and correct labels added a further 0.0325 (41.7\%; 13/15 wins, $p=0.00427$; Data Table T08). Supervised pretraining as such and the meaning of the labels thus both contribute. The reference for the source task is the released prediction, and the two shares apply to this protocol.

A handcrafted logistic-regression baseline matched to the same setting reached 0.5729 ($n=15$), only 0.0065 below the transfer arm; the full ladder of 18 methods is given in Data Table T14. The classical methods were evaluated only at this dense-window setting. At sparse hops, where the transfer gain is largest, the comparison between scratch and transfer is made within the same architecture.

\subsection{Replication on the Wearable dataset}
On the Wearable dataset, the normalization source again changed the transfer gain, but in the opposite direction (Table~\ref{tab:wearable}, Fig.~\ref{fig:atlas}c). With global-train normalization, transfer improved on scratch by only 0.0188 (0.5128 to 0.5317; 19/26 wins, Holm $p=0.3763$). With per-subject-z, the gain rose to 0.0569 (0.4903 to 0.5471; 19/26, Holm $p=0.02051$); the robust and baseline-referenced variants gave similar gains of 0.0568 and 0.0426, whereas differencing removed it ($-$0.0003). On WESAD, per-user normalization had shrunk the gain; on Wearable, it enlarged it. What the two datasets have in common is that the gain depends on how the input is normalized, and the direction of that dependence may reflect differences in their amplitude distributions. The hop scan, run under global-train normalization, showed a clear gain only at hop 240 (0.4377 to 0.5356, $\Delta=+0.0979$; 22/26, $p=7.02\times10^{-5}$), compared with 0.0139 and 0.0188 at hops 4 and 60 (Table~\ref{tab:wearablehop}, Fig.~\ref{fig:atlas}d).

\begin{table}[t]
\caption{Wearable normalization ($n=26$; Data Table T05). S/T: scratch/transfer; robust: per-subject robust; baseline-ref.: baseline-referenced.}
\label{tab:wearable}\centering\scriptsize
\setlength{\tabcolsep}{1.5pt}
\begin{tabular}{@{}lrrrrrrr@{}}
\toprule Normalization & S & T & $\Delta$ & Wins & $d_z$ & 95\% CI & Holm $p$\\
\midrule Global-train & 0.5128 & 0.5317 & +0.0188 & 19/26 & 0.251 & [-0.0099,0.0467] & 0.3763\\
Per-subject-z & 0.4903 & 0.5471 & +0.0569 & 19/26 & 0.641 & [0.0239,0.0912] & 0.0205\\
Subject robust & 0.4550 & 0.5119 & +0.0568 & 17/26 & 0.629 & [0.0221,0.0903] & 0.0234\\
Baseline-ref. & 0.4744 & 0.5170 & +0.0426 & 21/26 & 0.347 & [-0.0069,0.0863] & 0.0307\\
Differenced & 0.5273 & 0.5271 & $-$ 0.0003 & 14/26 & $-0.004$ & [-0.0242,0.0225] & 0.9403\\
\bottomrule
\end{tabular}
\end{table}

\begin{table}[t]
\caption{Wearable hop ladder (Data Table T06; $n=26$). Paired $p$ and $d_z$; window counts are given in Data Table T06.}
\label{tab:wearablehop}\centering\scriptsize
\setlength{\tabcolsep}{1.8pt}
\begin{tabular}{@{}rrrrrrrr@{}}
\toprule Hop & Overlap\% & Scratch & Transfer & $\Delta$ & Wins & $p$ & $d_z$\\
\midrule
4 & 98.3 & 0.5188 & 0.5327 & +0.0139 & 16/26 & 0.1430 & 0.300\\
60 & 75.0 & 0.5128 & 0.5317 & +0.0188 & 19/26 & 0.0941 & 0.251\\
240 & 0.0 & 0.4377 & 0.5356 & +0.0979 & 22/26 & $7.0\times10^{-5}$ & 0.876\\
\bottomrule
\end{tabular}
\end{table}

The two Wearable scans answer narrower questions than the WESAD grid. The normalization ladder holds the hop fixed and the hop scan holds the normalization fixed; the combinations in between were not run, so each scan supports comparisons only within its own rows. In the baseline-referenced row, the bootstrap interval for the mean difference includes zero while the corrected signed-rank test is significant. The two statistics describe different features of the paired distribution, and we report both alongside the prespecified test.

\subsection{Class-Level Effects, Probability Calibration, and Transition Windows}
The aggregate gains hide differences between classes. Under global-train normalization, transfer changed the accuracy for WESAD amusement by $-5.94$ percentage points, but that for baseline by $+11.96$ and for stress by $+9.80$; under per-subject-z, the largest net change was in amusement (+4.57 points; Data Table T10). A positive overall gain can therefore coexist with a loss in one class, and how the gains are distributed across classes shifts with the normalization; these net changes are counted over paired windows. Probability calibration behaved differently from discrimination. Subject-level expected calibration error changed little under per-subject-z (0.2116 to 0.2031, $\Delta=-0.0085$, $p=0.48871$) but decreased under differencing (0.2583 to 0.2263, $\Delta=-0.0320$, $p=0.00537$; Data Table T11), so gains in macro-F1 should not be read as gains in calibration. Excluding transition windows, with margins of 60 or 240 samples (15 or 60 s at 4 Hz), improved accuracy in both arms, by 0.0033 and 0.0040 and by 0.0152 and 0.0187 for scratch and transfer, respectively (Data Table T16); these summaries are not tied to a particular cell of the grid. Equivalence summaries and epoch curves are given in Data Tables T15 and T13.

\subsection{Signal-level diagnostics}
Fig.~\ref{fig:mechanism} shows what the decomposition branch computes on real windows and why its confidence needs an interpretation of its own. In the window with higher confidence, the tonic trace follows the input closely and the phasic and driver traces are zero. In the window with lower confidence, local departures from the smooth tonic component coincide with structure in the residual and with a drop in reconstruction confidence. No artifact annotations exist for these target windows, so the departures are deviations from the reconstruction and cannot be identified as artifacts. Both windows were chosen to show the extremes of confidence, and both agree with the high proportion of inactive drivers reported in Methods.

\section{Discussion}
The value of artifact-supervised pretraining cannot be stated without saying how the input was normalized. On WESAD, the gain over training from scratch ranged from 0.020 to 0.227, and with architecture and training budget held constant, the normalization source accounted for much of this range.

\subsection{Annotations and calibration as partially substitutable information}
Two results suggest that, on WESAD, artifact annotations and per-user calibration partly supply the same information: pretraining helped far more with artifact labels than with self-supervision on the same recordings, and per-user normalization improved the network trained from scratch more than the pretrained one at every hop. A simple account fits both observations. If pretraining has already made the network less sensitive to amplitude differences between people, per-user normalization leaves less for it to add. The substitution is partial, since the transfer arm kept positive point estimates under per-subject-z, and the lower sensitivity of the pretrained network to the hop was confined to the inductive setting (an across-hop SD ratio of 6.34 under global-train and 1.22 under per-subject-z). Amplitude invariance itself is a hypothesis consistent with these comparisons rather than a quantity we measured.

\subsection{First-use monitoring and protocol disclosure}
The largest gain occurred under global-train normalization, the only setting compatible with first use, whereas per-subject-z draws on the user's complete recording, including samples recorded after each prediction. Because signal amplitude and user characteristics vary widely in affect monitoring~\cite{vos2023generalizable,berkemeier2025affective,benchekroun2023crossdataset,albaladejo2022configurations}, this distinction is unlikely to be specific to our model. We therefore suggest that studies report the subject assignment, window length and hop, normalization formula, the data used for the normalization statistics, and when those data become available, items that fit existing reporting and risk-of-bias guidance~\cite{collins2024tripodai,collins2015tripod,wolff2019probast}. No particular formula is favored by this recommendation: a per-user baseline is legitimate when the intended workflow provides it, but it should be described by its acquisition interval, and complete-record normalization should be identified as retrospective.

\subsection{Scope and limitations}
The reversal on the Wearable dataset, where per-user normalization enlarged the gain, shows that the direction of the interaction depends on the data. Differences in tasks, amplitude distributions, and preprocessing may explain it, but the one-dimensional Wearable scans cannot separate these factors, and two identifiers missing from our processed copy reduced their power. Both downstream datasets record laboratory-evoked affect, and the 0.7815 wrist-only accuracy was obtained under per-user normalization, so neither result establishes performance in daily-life or clinical use. The fixed 12-epoch budget left losses at sparse hops unconverged, so part of the measured gain reflects faster optimization. The effect is also established for one backbone only: the ablations were not accompanied by paired component tests, the decomposition branch was mostly inactive, the classical baselines and the permutation control were run only at global-train, hop 1, and we make no claim of superiority over other compact models. Finally, the literature survey covers a fixed, non-exhaustive sample coded by a single author.

\section{Conclusion}
Artifact-supervised pretraining improved wrist-EDA affect recognition over training from scratch in 12 of 13 configurations across two datasets, but the size of the improvement depended on where the normalization statistics came from. On WESAD, the network pretrained on expert artifact labels gained 0.078 over scratch where self-supervised pretraining gained 0.008, and most of this advantage disappeared once the held-out user's own recording was used for normalization. Pretraining therefore mattered most in the situation that matters for first use, when a model must classify a new user without that user's statistics. On the Wearable dataset the effect of normalization went the other way, and the shuffled-label control shows that supervised pretraining as such contributes alongside the content of the labels.

For biomedical sensing, the practical consequence is that a reported score should state the split, window length, hop, normalization formula, source of the normalization statistics, and the time at which the user's data become available. Only five of the 50 WESAD studies we examined stated the source of their normalization statistics, so this gap is easy to close. An evaluation of first-use monitoring should use only information available at the time of prediction. Complete-record calibration answers a different question, and making clear which question a score answers is what allows published results to be compared.

\section*{Data and Ethics Statement}
EDABE, WESAD, and the PhysioNet Wearable dataset are public releases, accessed through their official distribution channels. EDABE refers to the 43-subject artifact-annotated release EDABE v2 (Mendeley Data, DOI 10.17632/w8fxrg4pv5.2) accompanying the artifact-recognition study~\cite{llanes2023edabe}. WESAD refers to the 2018 public release (DOI 10.1145/3242969.3242985)~\cite{schmidt2018wesad}. The Wearable cohort is drawn from PhysioNet v1.0.0 as defined in Methods. The accompanying data supplement provides Data Tables T01, T04--T08, and T10--T21 as CSV files, together with figure-generation code and source evidence. Reproducibility code and non-sensitive derived materials are available on GitHub (\url{https://github.com/rtb-1005/SAFE-EDA}) and Zenodo (\url{https://doi.org/10.5281/zenodo.22920792}); raw EDABE, WESAD, and PhysioNet data are not redistributed. No new participants were recruited. This study is a secondary analysis of public data, and ethics approvals for the original data collection are held by the respective dataset teams.

\section*{Conflict of Interest}
The authors declare no conflict of interest.

\section*{Author Contributions}
Haochen Chai led the study design, implementation, statistical analysis, manuscript preparation, and figure integration. Xinbi Luo contributed to data cleaning, data-structure checks, figure preparation, and literature retrieval and organization. Zining Liu contributed to data cleaning, table and figure verification, literature retrieval, and reference organization. Fangfang Jiang provided technical and funding support and reviewed the manuscript.

\section*{Acknowledgment}
The authors used Claude Opus (Anthropic) to polish the language of the manuscript and to assist with figure-drawing code; all content was reviewed by the authors, who take full responsibility for it.

\let\originalthebibliography\thebibliography
\renewcommand{\thebibliography}[1]{\originalthebibliography{#1}\fontsize{7pt}{7.7pt}\selectfont\setlength{\itemsep}{-0.6pt}\tolerance=1000\emergencystretch=2em}
\bibliographystyle{IEEEtran}
\begingroup\scriptsize\bibliography{ieee_style,refs}\endgroup
\end{document}


\bstctlcite{IEEEStyleControl}
\maketitle
\section*{Methods-text sample}
The sample comprises 50 published studies with available Methods text; it is a fixed, non-exhaustive sample coded by one author without an independent second coder.

Each study contributes once per field. Fields were coded only from Methods or experimental-setup text: a scaler name does not establish the statistics source, and a train/test ratio does not establish the allocation unit. Overlap stated explicitly is distinguished from overlap inferred from window and hop. An unresolved field means that the eligible text does not establish the information; it says nothing about whether the procedure itself was sound. Table~\ref{tab:survey50} identifies every study, and Table~\ref{tab:surveycounts} summarizes the counts.

The data supplement provides the full search strategy, DOI identities, coded values, model families, quotations, and PDF page locations. Data Tables T01, T04--T08, and T10--T21 contain all released experimental values as CSV files; paired inference uses seed-averaged subjects rather than windows or seeds. Definitions of diagnostic subsets and table-specific notes are recorded in the accompanying data dictionary.
\begin{strip}
\centering
\captionof{table}{Study-level Methods-text sample. W/H/O/S/L/A indicate whether window length, hop, explicit overlap, split type, LOSO, and metric aggregation are stated (1) or unresolved (--). Norm.: T = training data, U = evaluated-user window or baseline. References appear below; DOI identities and quotations are in the released study table.}
\label{tab:survey50}\centering\scriptsize
\setlength{\tabcolsep}{3pt}
\begin{tabular}{@{}lccccccc@{\hspace{12pt}}lccccccc@{\hspace{12pt}}lccccccc@{}}
\toprule Study & W & H & O & S & L & A & Norm. & Study & W & H & O & S & L & A & Norm. & Study & W & H & O & S & L & A & Norm.\\
\midrule
W001 \cite{surveyW001} & 1 & 1 & -- & 1 & 1 & -- & -- & W022 \cite{surveyW022} & -- & -- & -- & 1 & 1 & -- & -- & W042 \cite{surveyW042} & -- & -- & -- & -- & -- & -- & --\\
W002 \cite{surveyW002} & 1 & 1 & -- & 1 & 1 & 1 & -- & W023 \cite{surveyW023} & 1 & 1 & -- & 1 & 1 & -- & -- & W043 \cite{surveyW043} & 1 & 1 & 1 & 1 & 1 & 1 & --\\
W003 \cite{surveyW003} & 1 & -- & 1 & 1 & 1 & 1 & -- & W024 \cite{surveyW024} & 1 & -- & -- & -- & -- & -- & -- & W047 \cite{surveyW047} & -- & -- & -- & 1 & -- & -- & --\\
W004 \cite{surveyW004} & 1 & 1 & 1 & 1 & 1 & -- & -- & W025 \cite{surveyW025} & 1 & -- & -- & 1 & -- & -- & T & W048 \cite{surveyW048} & 1 & 1 & -- & -- & -- & -- & --\\
W005 \cite{surveyW005} & 1 & 1 & -- & 1 & 1 & -- & -- & W026 \cite{surveyW026} & 1 & -- & -- & 1 & -- & -- & -- & W049 \cite{surveyW049} & 1 & 1 & -- & -- & -- & -- & --\\
W007 \cite{surveyW007} & -- & -- & -- & -- & -- & -- & -- & W028 \cite{surveyW028} & -- & -- & -- & -- & -- & -- & -- & W050 \cite{surveyW050} & -- & -- & -- & 1 & -- & -- & --\\
W008 \cite{surveyW008} & -- & -- & -- & 1 & -- & -- & -- & W029 \cite{surveyW029} & -- & -- & -- & -- & -- & 1 & -- & W051 \cite{surveyW051} & -- & -- & -- & -- & -- & -- & --\\
W009 \cite{surveyW009} & -- & -- & -- & -- & -- & -- & -- & W030 \cite{surveyW030} & 1 & 1 & -- & -- & -- & -- & -- & W052 \cite{surveyW052} & 1 & -- & 1 & 1 & -- & 1 & U\\
W010 \cite{surveyW010} & 1 & 1 & -- & 1 & 1 & 1 & -- & W032 \cite{surveyW032} & -- & -- & -- & -- & -- & -- & -- & W053 \cite{surveyW053} & 1 & -- & -- & 1 & 1 & 1 & --\\
W011 \cite{surveyW011} & 1 & 1 & -- & 1 & -- & -- & U & W033 \cite{surveyW033} & -- & -- & -- & 1 & -- & 1 & -- & W054 \cite{surveyW054} & 1 & -- & -- & 1 & 1 & -- & --\\
W012 \cite{surveyW012} & 1 & 1 & -- & -- & -- & -- & -- & W034 \cite{surveyW034} & 1 & 1 & -- & 1 & -- & 1 & -- & W055 \cite{surveyW055} & -- & -- & -- & -- & -- & -- & --\\
W014 \cite{surveyW014} & -- & -- & -- & -- & -- & -- & -- & W035 \cite{surveyW035} & -- & -- & 1 & 1 & -- & -- & -- & W056 \cite{surveyW056} & -- & -- & -- & -- & -- & -- & --\\
W015 \cite{surveyW015} & -- & -- & -- & -- & -- & -- & -- & W036 \cite{surveyW036} & -- & -- & -- & -- & -- & -- & -- & W057 \cite{surveyW057} & -- & -- & -- & -- & -- & -- & --\\
W016 \cite{surveyW016} & -- & -- & -- & -- & -- & -- & -- & W037 \cite{surveyW037} & 1 & -- & -- & 1 & -- & -- & -- & W058 \cite{surveyW058} & -- & -- & -- & 1 & -- & -- & --\\
W017 \cite{surveyW017} & 1 & 1 & -- & 1 & -- & -- & U & W038 \cite{surveyW038} & 1 & -- & -- & 1 & -- & -- & -- & W059 \cite{surveyW059} & -- & -- & -- & -- & -- & -- & --\\
W020 \cite{surveyW020} & -- & -- & -- & 1 & -- & -- & -- & W039 \cite{surveyW039} & 1 & 1 & -- & 1 & 1 & -- & -- & W060 \cite{surveyW060} & -- & -- & -- & -- & -- & -- & --\\
W021 \cite{surveyW021} & 1 & -- & 1 & -- & -- & 1 & -- & W041 \cite{surveyW041} & 1 & -- & -- & 1 & 1 & 1 & U &  &  &  &  &  &  &  & \\
\bottomrule
\end{tabular}
\end{strip}

\begin{table}[t]
\caption{Methods-text reporting among 50 studies. Counts describe stated fields, not implementation quality. Intervals are 95\% Wilson score intervals; dashes indicate no computed interval.}
\label{tab:surveycounts}\centering\scriptsize
\begin{tabular}{@{}lrrl@{}}
\toprule Field & Studies & Percent & 95\% CI (\%)\\
\midrule Window length & 26/50 & 52 & [39,65]\\
Window hop & 15/50 & 30 & [19,44]\\
Explicit overlap & 6/50 & 12 & [6,24]\\
Split type & 28/50 & 56 & [42,69]\\
LOSO described & 13/50 & 26 & [16,40]\\
Metric aggregation & 11/50 & 22 & --\\
Normalization statistics source & 5/50 & 10 & [4,21]\\
Evaluated-user statistics used & 4/50 & 8 & --\\
\bottomrule
\end{tabular}
\end{table}
\FloatBarrier
\let\originalthebibliography\thebibliography
\renewcommand{\thebibliography}[1]{\originalthebibliography{#1}\fontsize{7pt}{7.7pt}\selectfont\setlength{\itemsep}{-0.6pt}}
\bibliographystyle{IEEEtran}
\begingroup\scriptsize\bibliography{ieee_style,survey_refs}\endgroup